\documentclass[letterpaper]{article} % DO NOT CHANGE THIS
\usepackage{aaai2026}  % DO NOT CHANGE THIS
\usepackage{times}  % DO NOT CHANGE THIS
\usepackage{helvet}  % DO NOT CHANGE THIS
\usepackage{courier}  % DO NOT CHANGE THIS
\usepackage[hyphens]{url}  % DO NOT CHANGE THIS
\usepackage{graphicx} % DO NOT CHANGE THIS
\usepackage{natbib}  % DO NOT CHANGE THIS AND DO NOT ADD ANY OPTIONS TO IT
\usepackage{caption} % DO NOT CHANGE THIS AND DO NOT ADD ANY OPTIONS TO IT
\usepackage{algorithm}
\usepackage{algorithmic}
\usepackage{subcaption}
\usepackage{booktabs} 
\usepackage{multirow}
\usepackage{newfloat}
\usepackage{listings}
\usepackage{amsmath} 
\usepackage{amssymb}   % 提供 \mathbb

\newtheorem{obs}{Observation}
\newtheorem{hyp}{Hypothesis}

\DeclareCaptionStyle{ruled}{labelfont=normalfont,labelsep=colon,strut=off} % DO NOT CHANGE THIS
\floatstyle{ruled}
\newfloat{listing}{tb}{lst}{}
\floatname{listing}{Listing}
\title{Revisiting Fairness-aware Interactive Recommendation: Item Lifecycle as a Control Knob }
 
\author{
    Yun Lu\textsuperscript{\rm 1,2},
    Xiaoyu Shi\textsuperscript{\rm 1,2,}\thanks{Corresponding author.},
    Hong Xie\textsuperscript{\rm 3},
    Chongjun Xia\textsuperscript{\rm 1,2},
    Zhenhui Gong\textsuperscript{\rm 1,2},
    Mingsheng Shang\textsuperscript{\rm 1,2}
}
\affiliations{

    \textsuperscript{\rm 1}Chongqing Institute of Green and Intelligent Technology, Chinese Academy of Sciences.\\
    \textsuperscript{\rm 2}Chongqing School, University of Chinese Academy of Sciences.\\
    \textsuperscript{\rm 3}The First Affiliated Hospital, University of Science and Technology of China\\
}

\begin{document}

\maketitle

\begin{abstract}
 This paper revisits fairness-aware interactive recommendation (e.g., TikTok, KuaiShou) by introducing a novel control knob, i.e., the lifecycle of items. We make threefold contributions. First, we conduct a comprehensive empirical analysis and uncover that item lifecycles in short-video platforms follow a compressed three-phase pattern, i.e., rapid growth, transient stability, and sharp decay, which significantly deviates from the classical four-stage model (introduction, growth, maturity, decline). Second,  we introduce LHRL, a lifecycle-aware hierarchical reinforcement learning framework that dynamically harmonizes fairness and accuracy by leveraging phase-specific exposure dynamics. LHRL consists of two key components: (1) PhaseFormer, a lightweight encoder combining STL decomposition and attention mechanisms for robust phase detection; (2) a two-level HRL agent, where the high-level policy imposes phase-aware fairness constraints, and the low-level policy optimizes immediate user engagement. This decoupled optimization allows for effective reconciliation between long-term equity and short-term utility. Third, experiments on multiple real-world interactive recommendation datasets demonstrate that LHRL significantly improves both fairness and user engagement. Furthermore, the integration of lifecycle-aware rewards into existing RL-based models consistently yields performance gains, highlighting the generalizability and practical value of our approach.

\end{abstract}

\begin{links}
    \link{Code}{https://github.com/luyunstar/LHRL}
    \link{Datasets}{https://kuairec.com\\
                    https://kuairand.com}
    \link{Extended version}{https://aaai.org/example/extended-version}
\end{links}

\section{Introduction}
Fairness in recommender systems (RSs) has gained increasing attention, especially in addressing \textit{popularity bias}, where a few popular items dominate exposure while long-tail content is underrepresented. This imbalance is often reinforced by learning algorithms and further amplified in \textbf{interactive recommendation} settings (e.g., TikTok, KuaiShou), where feedback loops and time-varying user preferences exacerbate unfair exposure. Thus, providing equitable item exposure throughout interaction cycles, regardless of item popularity, becomes a fundamental requirement for an interactive recommender system (IRS).

Prior work has proposed various fairness-aware strategies, e.g., incorporating causal inference techniques to disentangle bias signals~\cite{zheng2021disentangling,wei2021model}, adding fairness-oriented regularization terms into training objectives~\cite{konstantinov2021fairness}, and applying re-ranking algorithms to improve post-hoc exposure balance~\cite{abdollahpouri2019managing,zhu2021popularity}. However, most of them focus on static or one-shot recommendation settings, overlooking the dynamic nature of item popularity. Even RL-based methods~\cite{r-19,sac,r-23} (capable of sequential decision-making) largely ignore a key temporal factor: the item lifecycle. In practice, items go through temporal phases of exposure demand, which are rarely uniform or static. Ignoring these phases can lead to misaligned exposure. For instance, stale content may be over-promoted, while emerging items miss critical early exposure opportunities.
\begin{figure}[!t]
  \centering
  \includegraphics[width=0.48\textwidth]{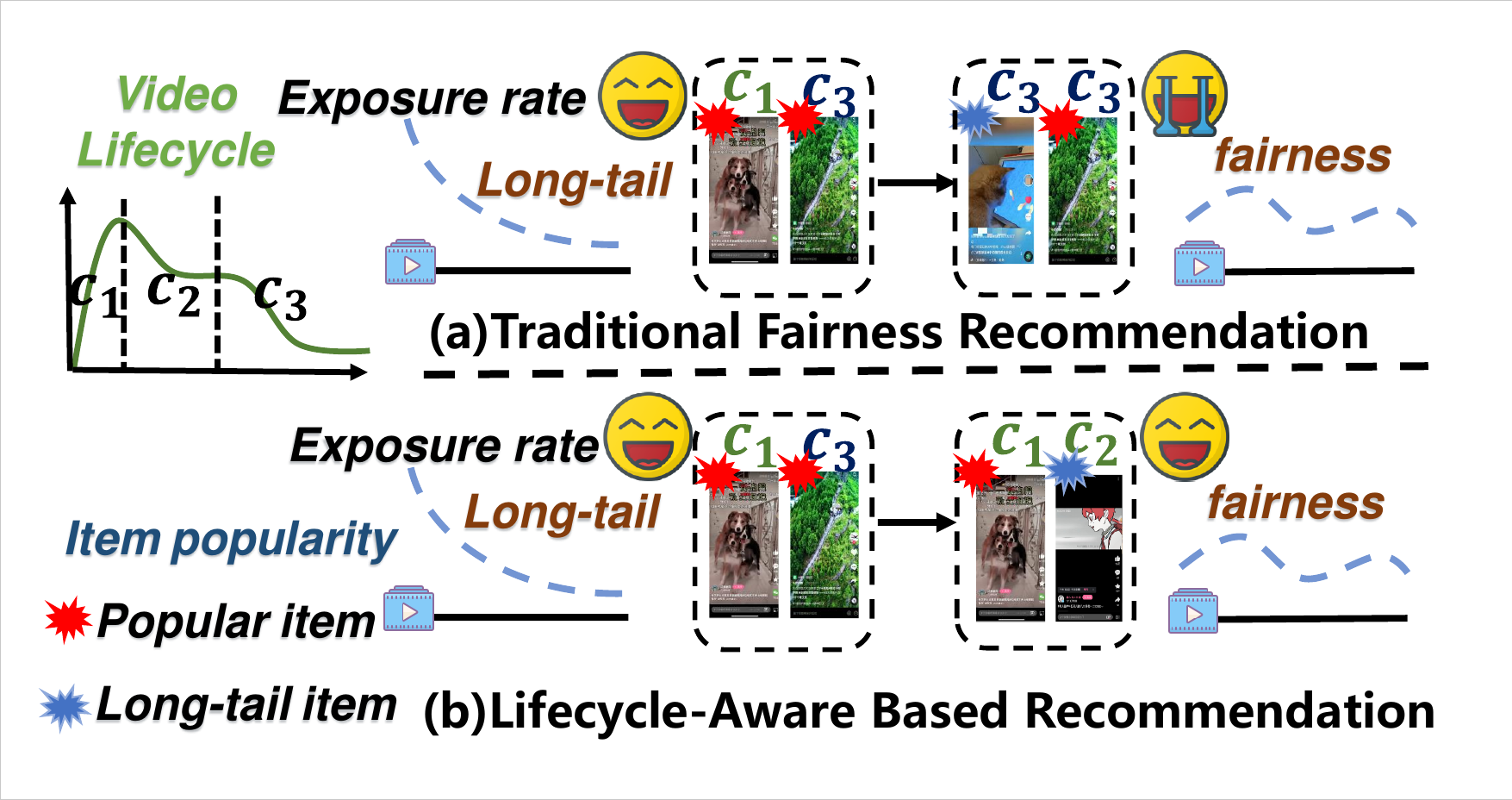}
  \caption{The role of lifecycle in fairness recommendation}
  \label{fig:lifecycle}
 \end{figure}

To further investigate this overlooked factor, we conduct an in-depth analysis of item exposure patterns in real-world short-video platforms (e.g., KuaiShou). Interestingly, we observe that short-video content often follows a distinct lifecycle trajectory that departs from the classic long-tail assumption. As shown in Figure~\ref{fig:lifecycle}, items typically undergo multiple temporal phases: rapid growth (C1), followed by transient stability (C2), and sharp decay (C3). Unfortunately, existing fairness-aware recommendation methods (whether static or RL-based) often ignore such lifecycle-aware dynamics. They tend to apply uniform constraints across all items, regardless of their lifecycle state, which may result in mismatches between exposure demand and exposure allocation. As illustrated in Figure~\ref{fig:lifecycle}, conventional fairness strategies often apply uniform constraints across all items, ignoring the temporal heterogeneity in content demand. For instance, static fairness methods tend to over-promote stale content in the decay phase (C3), where user interest has diminished, leading to wasted exposure opportunities. Meanwhile, reinforcement learning methods, though adaptive, still lack lifecycle awareness, resulting in insufficient exposure for emerging content during the critical growth (C1) and stability (C2) stages. In contrast, our proposed method dynamically adjusts exposure based on lifecycle phase, allocating visibility where it is most impactful: boosting new items early, sustaining popular items briefly, and gracefully fading obsolete ones. This phase-aware strategy harmonizes fairness and efficiency over time. Hence, we argue that explicitly modeling the item lifecycle can offer a finer-grained lens to guide fair exposure decisions. By adapting recommendations to the temporal needs of each item, the system can mitigate entrenched popularity bias while maintaining user satisfaction.

Despite its promise, implementing lifecycle-aware fair recommendation raises two fundamental challenges: (1) \textit{How can we automatically uncover the latent lifecycle patterns of items?} (2) \textit{How can we dynamically adjust fairness strategies according to lifecycle phases?}

To address the above challenges, we propose LHRL, a Lifecycle-aware Hierarchical Reinforcement Learning framework, for interactive recommendation. Specifically, we design PhaseFormer, a lightweight encoder that leverages STL decomposition and attention-based modeling for robust lifecycle phase detection. It integrates STL decomposition to extract trend/seasonal/residual components, and employs iTransformer to encode them as tokens, enabling phase recognition through attention-based dependency modeling. Second, we introduce a hierarchical reinforcement learning agent that decouples long-term fairness from short-term engagement. The high-level policy adapts fairness constraints based on phase-popularity distributions via Gaussian sampling, while the low-level policy refines exposure decisions under these constraints, balancing the promotion of emerging content and the suppression of stale items. Together, LHRL enables phase-aware exposure strategies that dynamically harmonize fairness and accuracy over time. Our main contributions are summarized as follows:
\begin{itemize}

\item We identify item lifecycle as a key but underexplored control knob for fairness in interactive recommendation, revealing distinct multi-phase exposure patterns beyond the long-tail assumption.

\item We propose LHRL, a lifecycle-aware hierarchical reinforcement learning framework that dynamically aligns exposure with lifecycle stages via signal decomposition and phase-guided policy optimization.

\item We conduct extensive experiments on real-world datasets to verify the effectiveness and generality of LHRL in improving both fairness and accuracy, with plug-and-play compatibility for existing RL-based methods.
\end{itemize}

\section{Empirical Study On Short-Video Lifecycle}
\begin{figure}[!t]
\centering
\begin{subfigure}[t]{.48\linewidth}
  \includegraphics[width=\linewidth]{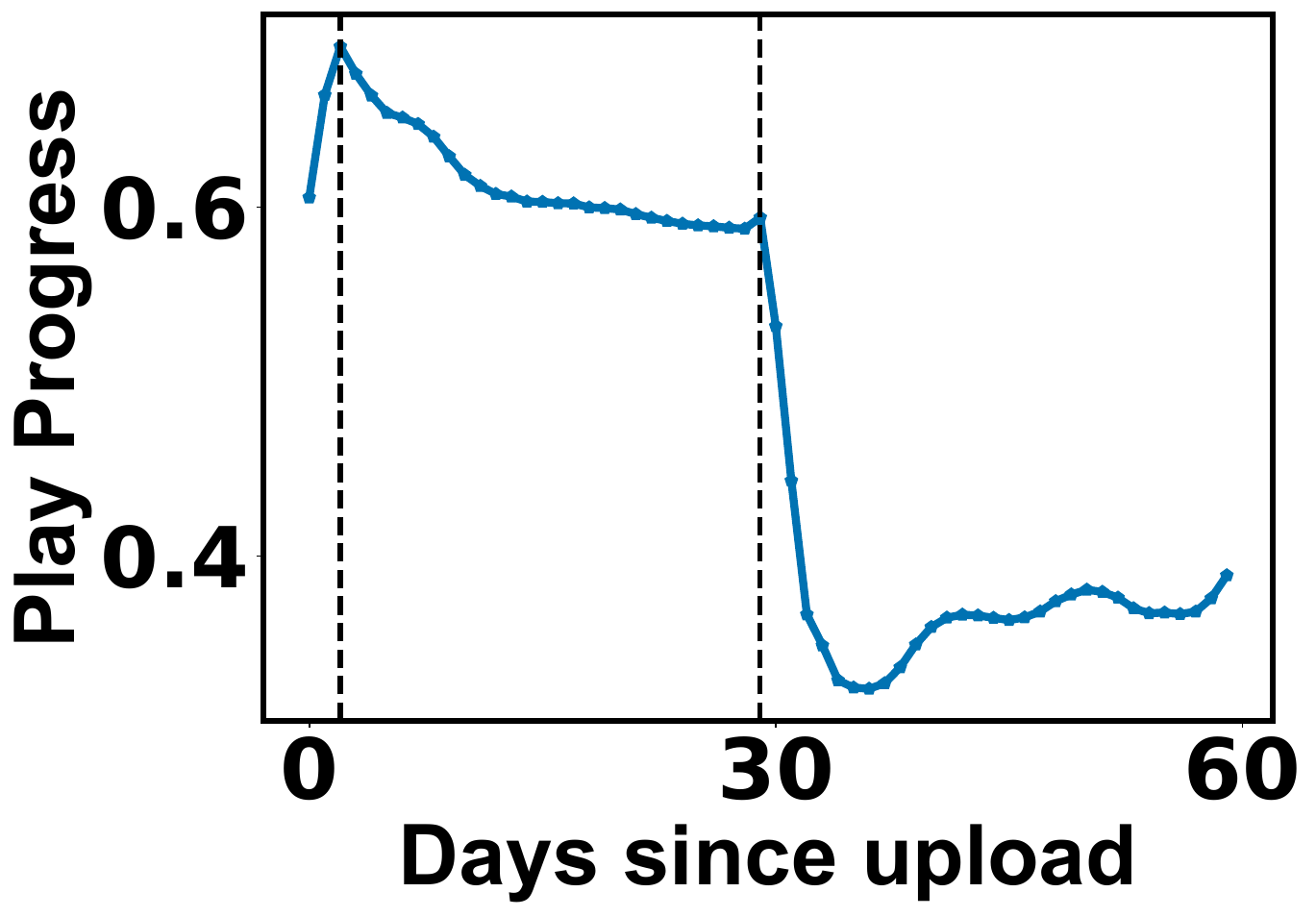}
  \caption{KuaiRec}\label{life:rec}
\end{subfigure}\hfill
\begin{subfigure}[t]{.48\linewidth}
  \includegraphics[width=\linewidth]{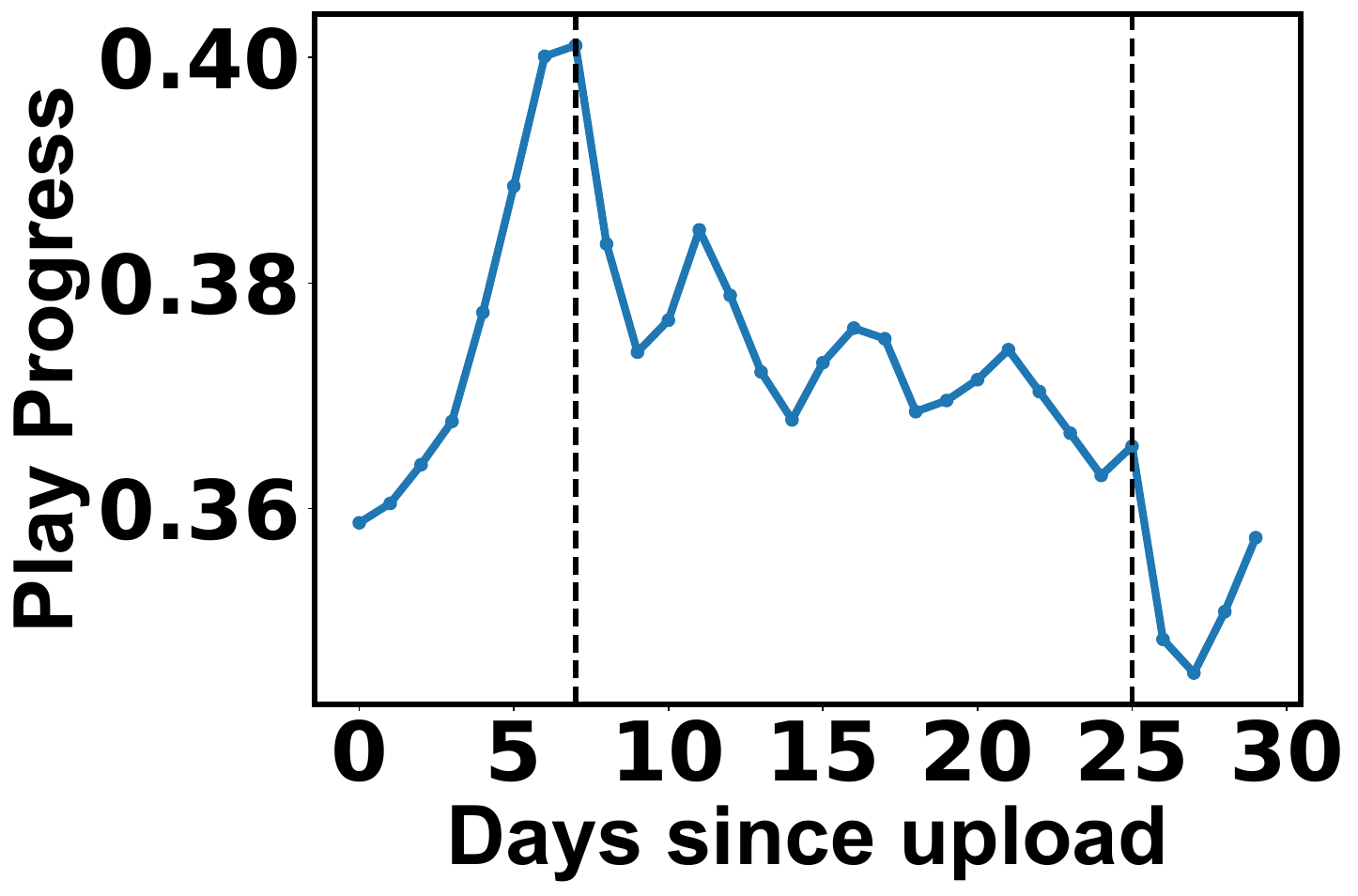}
  \caption{KuaiRand}\label{life:rand}
\end{subfigure}

\begin{subfigure}[t]{.48\linewidth}
  \includegraphics[width=\linewidth]{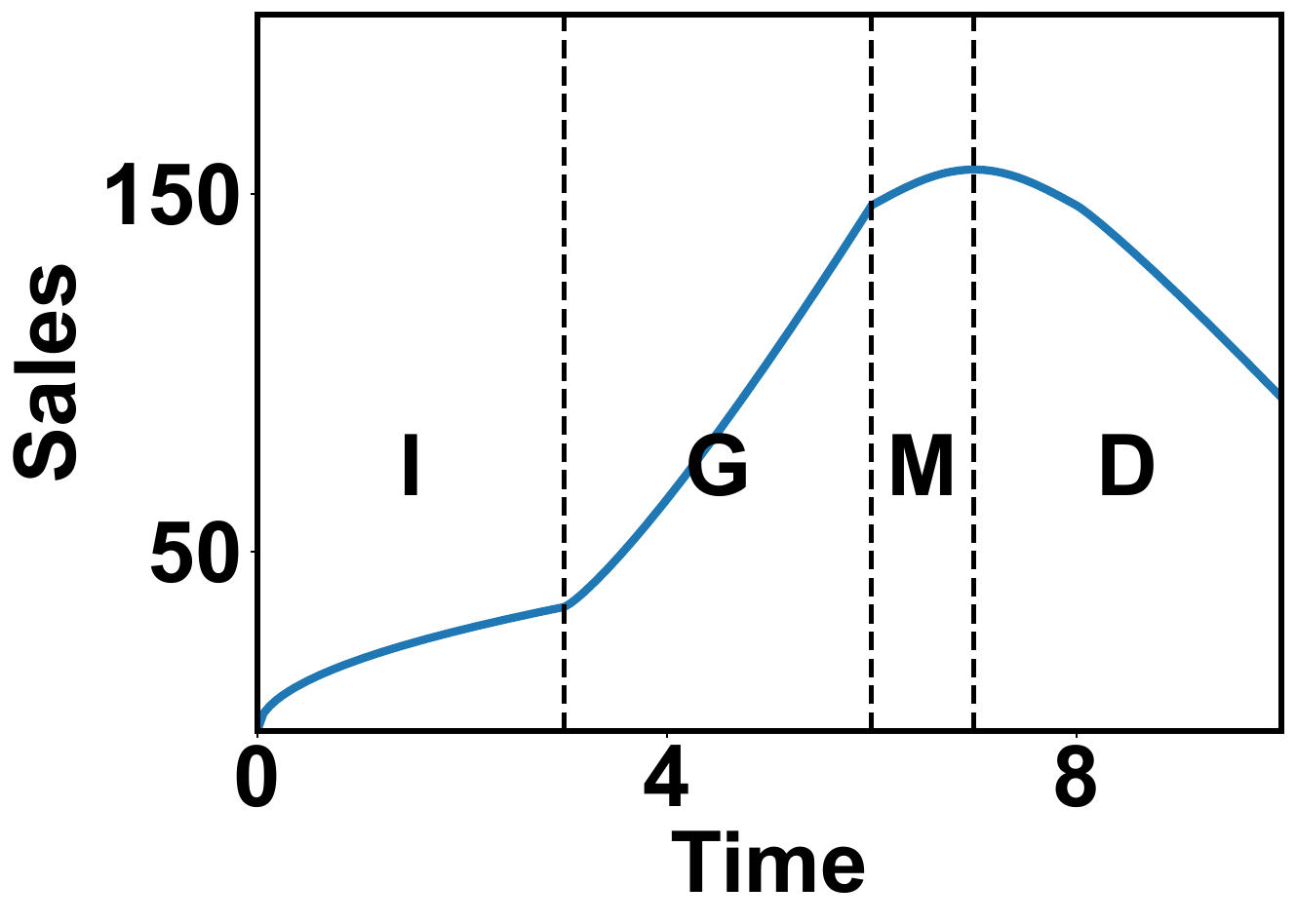}
  \caption{Product Lifecycle}\label{old:1}
\end{subfigure}\hfill
\begin{subfigure}[t]{.48\linewidth}
  \includegraphics[width=\linewidth]{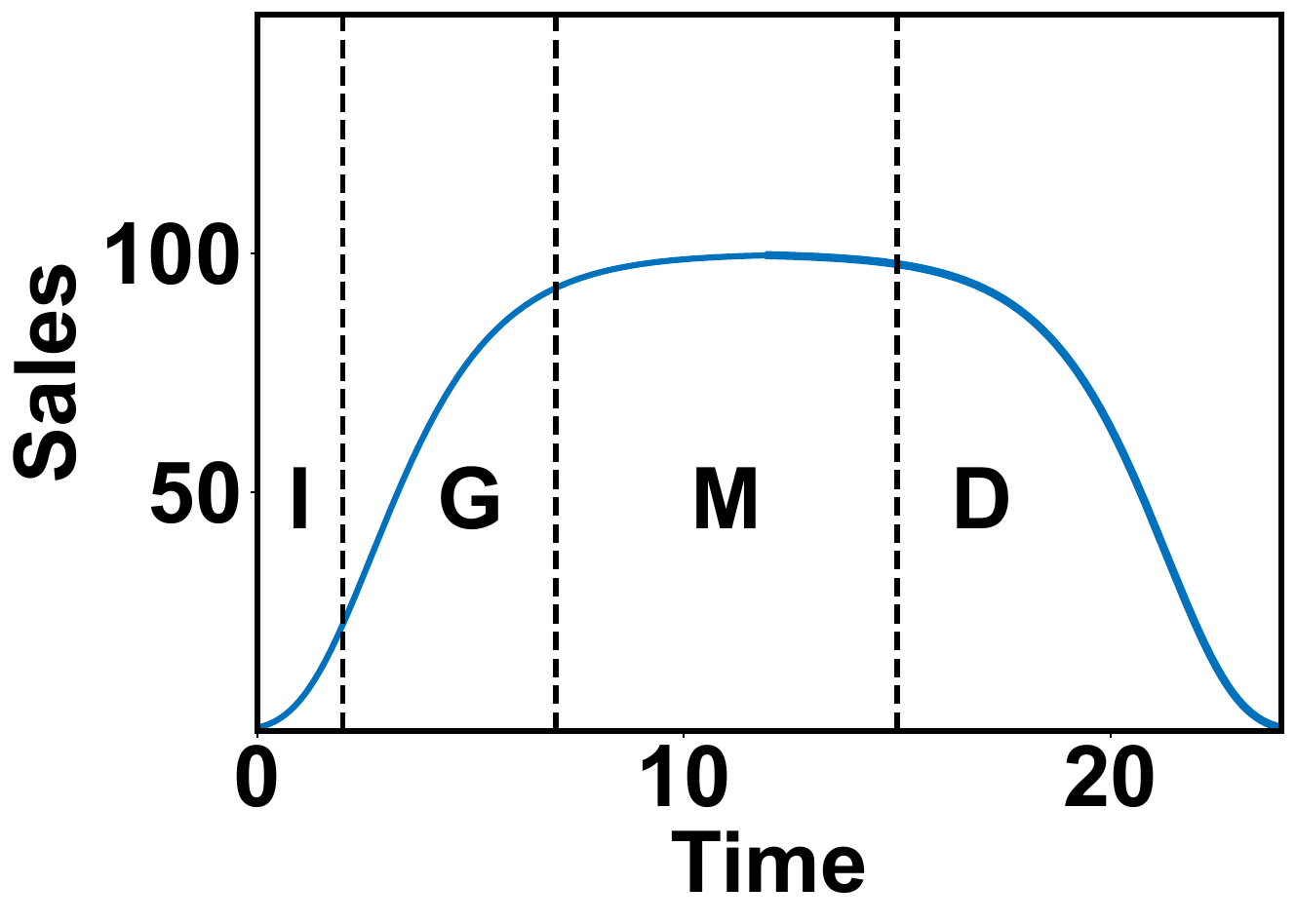}
  \caption{Gompertz Curve}\label{old:2}
\end{subfigure}

\begin{subfigure}[t]{.48\linewidth}
  \includegraphics[width=\linewidth]{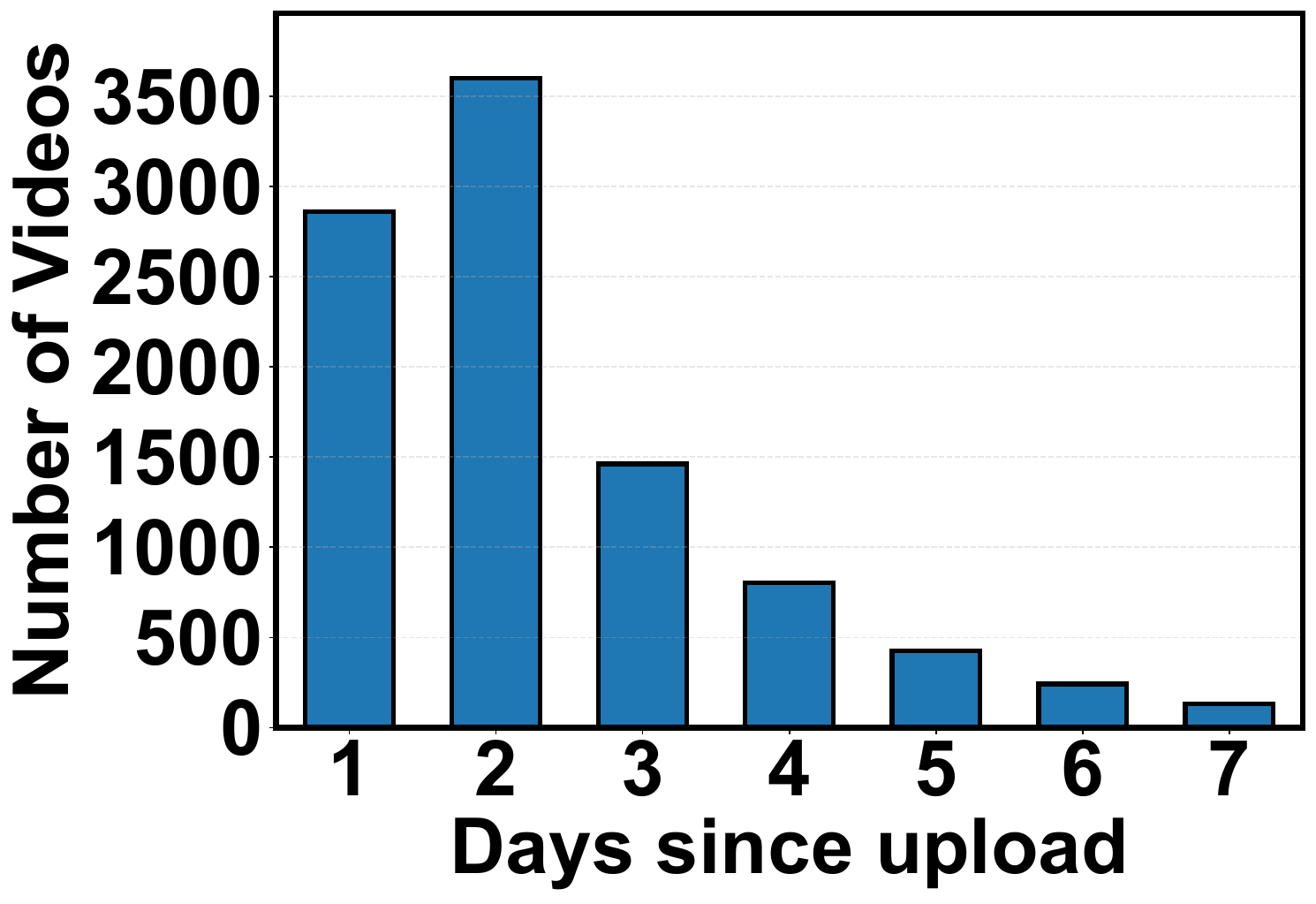}
  \caption{KuaiRec}\label{growth:rec}
\end{subfigure}\hfill
\begin{subfigure}[t]{.48\linewidth}
  \includegraphics[width=\linewidth]{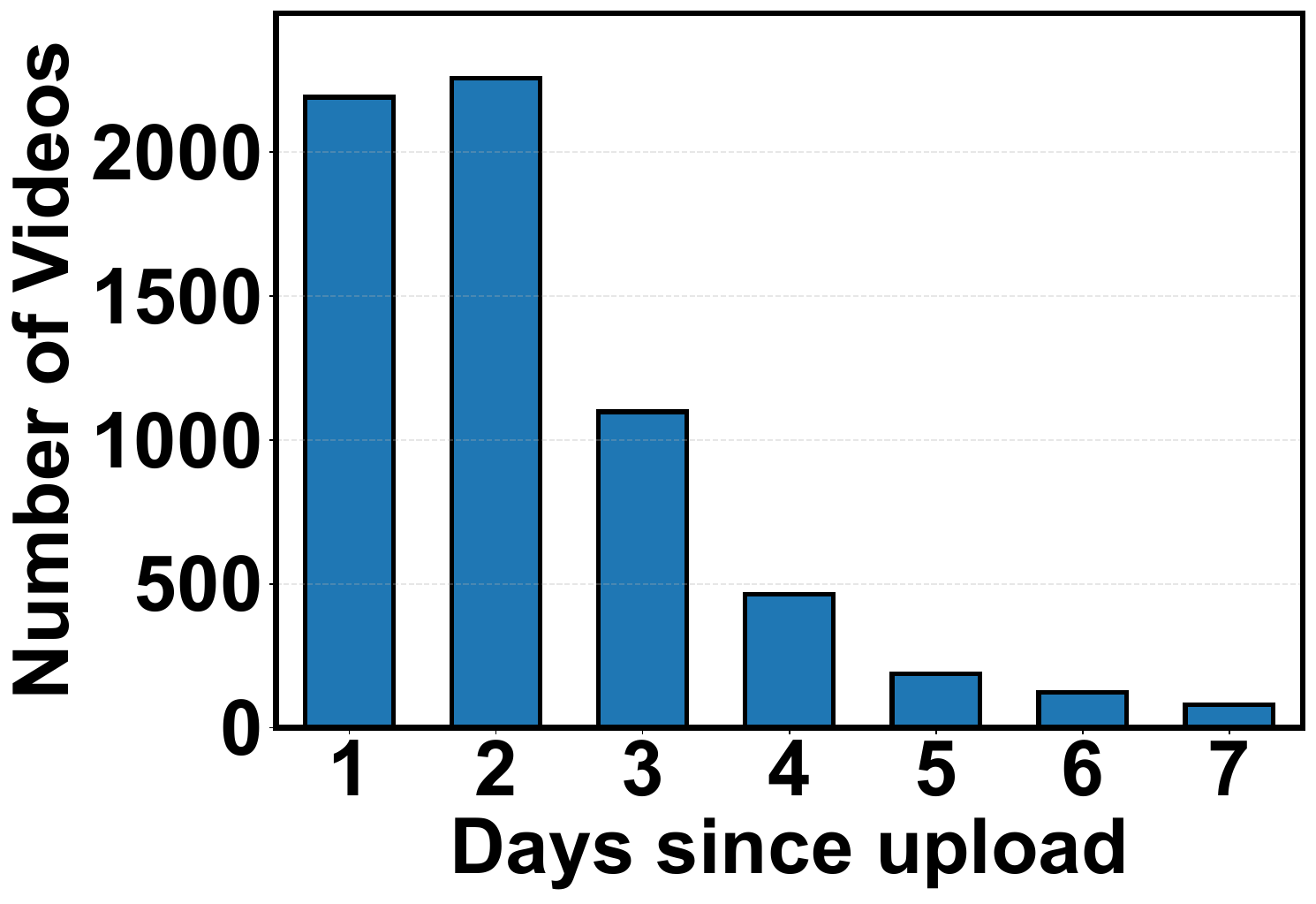}
  \caption{KuaiRand}\label{growth:rand}
\end{subfigure}
\caption{%
  Lifecycle analysis on KuaiRec and KuaiRand:
 (a)--(b) aggregated play progress curves; 
(c) classical four-phase product lifecycle (PLC); 
(d) Gompertz growth curve; 
(e)--(f) distribution of growth-phase duration.
  92.9\,\% (9524/10253, KuaiRec) and 94.5\,\% (6400/6777, KuaiRand) of videos reach peak engagement within the first 7 days.
}
\label{fig_all}
\end{figure}

% To investigate the intrinsic temporal dynamics of short videos, we conduct a comprehensive empirical analysis on two real-world datasets, KuaiRec\footnote{https://kuairec.com/}~\cite{rec} and KuaiRand\footnote{https://kuairand.com/}~\cite{rand}. For KuaiRand, we analyze 331,563 play records from 10,666 videos (after excluding 413 videos with no engagement), which collect interactions from April 10, 2022, to May 8, 2022. For KuaiRec, we process 102,661 play records from 7,262 videos, similarly filtering out 485 invalid entries to ensure data integrity, which spans the period from July 5, 2020, to September 5, 2020.

To investigate the intrinsic temporal dynamics of short videos, we conduct a comprehensive empirical analysis on two real-world datasets, KuaiRec~\cite{rec} and KuaiRand~\cite{rand}. For KuaiRand, we analyze 331,563 play records from 10,666 videos (after excluding 413 videos with no engagement), which collect interactions from April 10, 2022, to May 8, 2022. For KuaiRec, we process 102,661 play records from 7,262 videos, similarly filtering out 485 invalid entries to ensure data integrity, which spans the period from July 5, 2020, to September 5, 2020.

We define video vitality by \textbf{play progress}, defined as the ratio of play duration to video duration, thereby mitigating bias from video length and accidental clicks. Comparing these dynamics with classical product lifecycle models~\cite{draskovic2018product} (see Figure~\ref{old:1}, \ref{old:2}), we observe two major mismatches: 1) Short videos often attract attention immediately upon release, lacking a distinct "introduction" stage. As shown in Figures~\ref{growth:rec} and \ref{growth:rand}, more than 90\% of videos reach peak activity within 1–7 days. 2) The lifecycle exhibits pronounced asymmetry, i.e., rapid initial growth is followed by an extended and sharp decline in engagement. These findings motivate the following observation:
\begin{obs} \label{obs:1}
    Short videos follow a compressed three-stage lifecycle.
\end{obs} 

Based on the aggregated play progress curves (see Figures~\ref{life:rec} and \ref{life:rand}), we define the following three lifecycle stages:
\begin{itemize}
    \item \textbf{Growth Phase}: Marked by a rapid increase in play progress immediately after release.
    \item \textbf{Mature Phase}: A period of relative stability or gradual decline in play progress following its peak.
    \item \textbf{Decline Phase}: A stage of steep and accelerated decrease in user engagement.
\end{itemize}

We identify phase boundaries through differential analysis on the play progress curve, locating the peak and its subsequent inflection point. Thus, we get the following observation:
\begin{obs}
Lifecycle stages are statistically distinguishable and encode intrinsic item value.
\end{obs}

\begin{table}[!t]
\centering
\small
\resizebox{0.48\textwidth}{!}{
\begin{tabular}{llcccc}
\toprule
\multirow{2}{*}{Dataset} & \multirow{2}{*}{Group} & \multirow{2}{*}{Comparison} & \multirow{2}{*}{$p$ (adj.)} & \multirow{2}{*}{Sig.} & Mean \\
 & & & & & diff. \\
\midrule
\multirow{6}{*}{KuaiRec}
 & \multirow{3}{*}{Popular}
 & Growth vs.\ Mature & 0.418551 & ns & 7.8\% \\
 & & Growth vs.\ Decline & 5.41e-05 & *** & 43.3\% \\
 & & Mature vs.\ Decline & 6.00e-12 & *** & 38.4\% \\
 \cmidrule(lr){2-6}
 & \multirow{3}{*}{Long-tail}
 & Growth vs.\ Mature & 0.224092 & ns & 15.4\% \\
 & & Growth vs.\ Decline & 1.37e-05 & *** & 47.4\% \\
 & & Mature vs.\ Decline & 3.03e-11 & *** & 37.9\% \\
\midrule
\multirow{6}{*}{KuaiRand}
 & \multirow{3}{*}{Popular}
 & Growth vs.\ Mature  & 0.009596 & **  & 3.5\% \\
 & & Growth vs.\ Decline & 4.42e-04 & *** & 16.4\% \\
 & & Mature vs.\ Decline & 4.21e-06 & *** & 13.3\% \\
 \cmidrule(lr){2-6}
 & \multirow{3}{*}{Long-tail}
 & Growth vs.\ Mature  & 0.327009 & ns  & 2.5\% \\
 & & Growth vs.\ Decline & 0.001773 & **  & 16.7\% \\
 & & Mature vs.\ Decline & 3.97e-08 & *** & 17.1\% \\
\bottomrule
\end{tabular}%
}
\caption{Post-hoc comparisons (Dunn test with Holm--Bonferroni correction) across lifecycle stages. Mean diff.\ $>0$ indicates the first stage outperforms the second.}
\label{tab:lifecycle_comparison}
\end{table}

To validate these distinctions, we conduct Kruskal–Wallis tests (and ANOVA where applicable) across the three phases, confirming significant differences ($p<0.001$) in user engagement levels. Furthermore, we stratify videos into popular (top 20\% by exposure) and long-tail (bottom 80\%) groups. As shown in Table~\ref{tab:lifecycle_comparison}, the Decline phase shows statistically significant reductions in play progress ($p<0.01$), than both the Growth and Mature phases, up to 43.3\% and 38.4\% lower in KuaiRec popular videos, respectively. Notably, the average initial slope in the Decline phase is approximately 9.04 times that of the Mature phase, revealing the steepness of user disengagement. Conversely, play progress in Growth and Mature phases is statistically comparable ($p>0.05$), indicating similar user attractiveness during these stages.

These trends hold across datasets, popularity groups, and various video attributes (as detailed in the Appendix), confirming the generalizability of the proposed lifecycle model.

The above analysis leads us to the following hypothesis:
\begin{hyp}
An item's lifecycle stage is both predictive and reflective of its intrinsic value, and can serve as a dynamic control signal for recommendation. Accurately identifying this stage enables more informed exposure strategies.
\end{hyp}

 \begin{figure*}[!t]
  \centering
  \includegraphics[width=\textwidth]{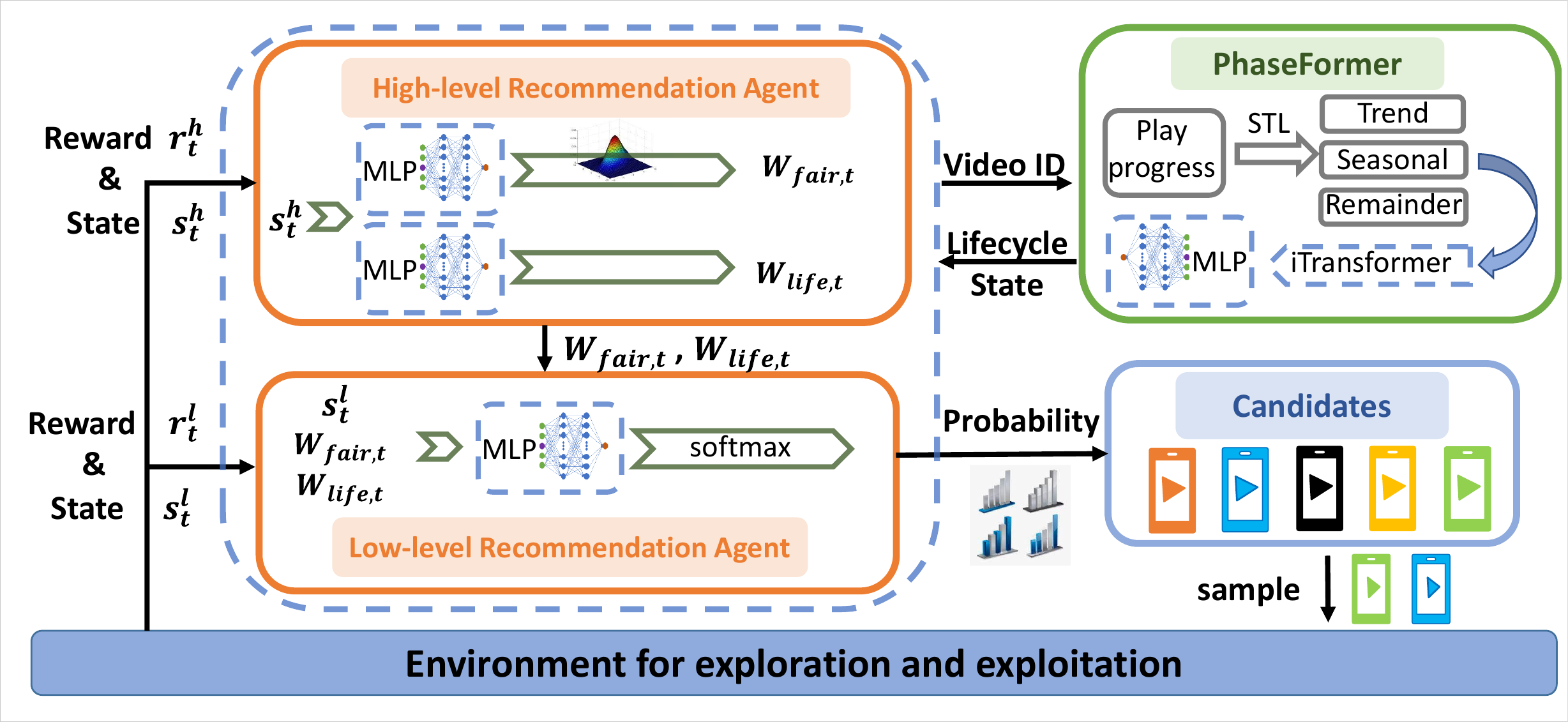}
  \caption{ The overall architecture of LHRL}
  \label{overall}
 \end{figure*}

\section{LHRL}
Our Lifecycle-aware Hierarchical Reinforcement Learning (LHRL) framework aims to dynamically coordinate fairness and accuracy within interactive recommender systems by leveraging item lifecycle stages as a pivotal control mechanism. LHRL comprises three main modules: the PhaseFormer module for predicting items' real-time lifecycle stages; the High-level Recommendation Agent (HRA) for setting stage-specific fairness constraints; and the Low-level Recommendation Agent (LRA) for optimizing immediate user engagement while adhering to HRA's guidance. An overview of the LHRL framework is presented in Figure~\ref{overall}.

\subsection{PhaseFormer Module}
The PhaseFormer module is designed to model the dynamic evolution of item popularity as a non-stationary latent process and infer the item's current lifecycle stage in real time.

We employ a two-stage representation learning strategy. First, we decompose the raw play progress time series $Y = \{y_1, y_2, ..., y_T\}$ using the Seasonal-Trend decomposition with Loess (STL):
\begin{equation}
Y_t = T_t + S_t + R_t,
\end{equation}
where $T_t$ denotes the long-term trend, $S_t$ captures seasonal patterns, and $R_t$ represents residual noise. Given the non-stationary and weakly periodic nature of short video engagement signals, STL offers robustness by orthogonally decoupling the temporal structure without prior distributional assumptions. This facilitates a stable and interpretable representation for downstream learning.

The extracted $T_t$ and $S_t$ components are then fed into an iTransformer encoder, which leverages global attention to build high-order, long-range representations. Compared to recurrent models (e.g., LSTM), this design mitigates the influence of local noise and enhances phase recognition robustness. The output is passed to a multi-layer perceptron (MLP) classifier to predict the current lifecycle stage among \texttt{[Growth, Mature, Decline]}:
\begin{equation}
P(\text{Stage}) = \phi\left(\text{MLP}\left(\text{iTransformer}\left(T_t, S_t\right)\right)\right),
\end{equation}
where $\phi(\cdot)$ denotes the softmax normalization.

The inferred lifecycle stage acts as a control signal guiding hierarchical decision-making in the RL framework. For new items with limited history, PhaseFormer uses masked self-attention and time-series extrapolation (e.g., linear projection of $T_t$) to handle sparse observations.

\subsection{Hierarchical Reinforcement Learning}
Our LHRL adopts a two-level hierarchical reinforcement learning design: the High-level Recommendation Agent (HRA) governs long-term fairness objectives, while the Low-level Recommendation Agent (LRA) focuses on short-term user engagement.

\textbf{State Representation.} We construct two user-specific state vectors: $\mathbf{s}_t^h$ for HRA and $\mathbf{s}_t^l$ for LRA. These are obtained via separate user encoders that extract interaction history and popularity preference patterns.

\textbf{High-level Recommendation Agent.} The HRA's primary responsibility is to dynamically generate fairness weights based on the user state and learn adaptive lifecycle weights to guide the LRA's item selection.

1) High-level Policy: The HRA generates two sets of weights: fairness weights $\mathbf{w}_{fair}$ and lifecycle weights $\mathbf{w}_{life}$. The fairness weights are sampled from a multivariate normal distribution parameterized by the output of an MLP:
\begin{equation}
\mathbf{w}_{fair, t} = [w_{pop}, w_{longtail}] \sim \mathcal{N}(\boldsymbol{\mu}_t^h, \sigma^2),
\end{equation}
where $\boldsymbol{\mu}_t^h = \text{MLP}_h(\mathbf{s}_t^h)$ and $\sigma$ is a learnable scalar. The lifecycle weights are directly predicted as:
\begin{equation}
\mathbf{w}_{life, t} = [w_{growth}, w_{mature}, w_{decline}] = \text{MLP}_{life}(\mathbf{s}_t^h).
\end{equation}
Both weight vectors are normalized via softmax and modulated using a temperature parameter $\tau$ to control their entropy.

2) High-level Reward: The high-level reward combines three components:
\begin{equation}
r_t^h = r_t^a + \alpha \cdot r_{life, t} + \beta \cdot r_{fair, t},
\label{eq:weight}
\end{equation}
where $r_t^a$ denotes click-based user feedback, $r_{life, t}$ encourages exposure to high-potential lifecycle stages, and $r_{fair, t}$ rewards long-tail item promotion.

The Lifecycle-aware reward is designed as:
\begin{equation}
r_{life, t} =  \mathbf{I}_{pop, t} \cdot \Lambda  \cdot  \mathbf{I}_{life, t}^T, \\
\end{equation}
where $\mathbf{I}_{pop,t} = [\mathbb{I}_{longtail}, \mathbb{I}_{popular}]$ indicates item popularity group, $\Lambda \in \mathbb{R}^{2 \times 3}$ is a reward coefficient matrix, and $\mathbf{I}_{life,t}$ is a one-hot vector representing the item's current stage.

The Fairness-aware reward is calculated by:
\begin{equation}
r_{fair, t} = \frac{\gamma_t^f}{1 + e_t^p},
\end{equation}
where $\gamma_t^f$ is the user's demand for fairness, and $e_t^p$ denotes their bias toward popular content.

HRA is trained using Proximal Policy Optimization (PPO)~\cite{ppo} with clipped surrogate loss.
% \begin{equation}
% \begin{aligned}
% \mathbb{E}_{t}\bigg[
% \min\bigg(
% &\frac{\pi_{\theta}(a_t|\mathbf{s}_t)}{\pi_{\theta_{old}}(a_t|\mathbf{s}_t)} \hat{A}_t, \\
% &\operatorname{clip}\left(\frac{\pi_{\theta}(a_t|\mathbf{s}_t)}{\pi_{\theta_{old}}(a_t|\mathbf{s}_t)}, 1-\epsilon, 1+\epsilon\right) \hat{A}_t
% \bigg)
% \bigg].
% \end{aligned}
% \label{eq:ppo}
% \end{equation}

\textbf{Low-level Recommendation Agent}. The LRA generates the final recommendation list under the guidance of fairness and lifecycle weights output by the HRA.
1) Low-level State: The LRA receives an enhanced low-level state that concatenates $\mathbf{s}_t^l$ with the high-level weights:
\begin{equation}
\tilde{\mathbf{s}}_t^l = \text{concat}(\mathbf{s}_t^l, \mathbf{w}_{fair, t}, \mathbf{w}_{life, t}).
\end{equation}

2) Low-level Policy: The LRA's policy is an MLP that takes the enhanced low-level state as input and outputs the raw interest scores $\mathbf{g}$ for all candidate items. The final item recommendation score is calculated by integrating user interest, item popularity, and HRA's fairness weights. The specific calculation process is as follows:
% \begin{equation}
% P(I \mid \tilde{\mathbf{s}}_t^l) = \phi \left( \left( \mathbf{g} \odot (\mathbf{I}_{pop,t} \cdot \mathbf{w}_{fair,t}) \odot (1+\mathbf{I}_{life,t} \cdot \mathbf{w}_{life,t}) \right) / \tau \right),
% \end{equation}
\begin{equation}
\begin{split}
P(I \mid \tilde{\mathbf{s}}_t^l) = \phi \bigg( & \left( \mathbf{g} \odot \left( \mathbf{I}_{pop,t} \cdot \mathbf{w}_{fair,t} \right) \right. \\ % <-- 在这里添加 \right.
     & \left. \odot \left( 1+\mathbf{I}_{life,t} \cdot \mathbf{w}_{life,t} \right) \right) / \tau \bigg), % <-- 在这里添加 \left.
\end{split}
\end{equation}
where $\mathbf{g} = \text{MLP}_l(\tilde{\mathbf{s}}_t^l)$ is the base relevance score. The final sampling distribution encourages the discovery of promising long-tail content while suppressing declining popular items.
The LRA's policy is also optimized through PPO, with an objective function similar to the HRA objective function.
% shown in Equation (\ref{eq:ppo}).

3) Low-level Reward: The LRA is optimized using PPO with the reward signal defined as the immediate interaction feedback:
\begin{equation}
r_t^l = r_t^a.
\end{equation}

This two-level design enables LHRL to simultaneously promote high-quality long-tail content and reduce overexposure of declining popular items, achieving a harmony between short-term user satisfaction and long-term ecosystem fairness.

\section{Experiments}
We conduct comprehensive experiments to address five research questions:
\begin{itemize}
  \item \textbf{RQ1:} How does LHRL compare to state-of-the-art methods in terms of long-term user satisfaction and fairness?
  \item \textbf{RQ2:} What are the individual contributions of LHRL's core components (hierarchy and lifecycle rewards)?
  \item \textbf{RQ3:} How accurately does PhaseFormer predict lifecycle stages, particularly under sparse early-stage data?
  \item \textbf{RQ4:} Can lifecycle-aware rewards generalize to other reinforcement learning frameworks?
  \item \textbf{RQ5:} What adaptive strategies does LHRL learn to manage exposure across lifecycle stages dynamically?
\end{itemize}
To assess the practicality and robustness of our method, we include two additional studies in Appendix: 
1) a time-efficiency analysis showing that the lifecycle module introduces only moderate training overhead, and 
2) a hyperparameter evaluation about the lifecycle reward weight ($\alpha$) and fairness reward weight ($\beta$), confirming that performance is stable when hyperparameters are appropriately emphasized. 

\subsection{Experiment Setup}
\textbf{Recommendation Environments}. Experiments are conducted in KuaiSim, an interactive simulator based on the KuaiRec and KuaiRand datasets. It models sessions of up to 20 interactions, featuring early termination for recommendation unsatisfaction or unfairness, and provides user feedback via a pre-trained DeepFM model. Our evaluation protocol is two-fold: online RL agents are trained through direct interaction, while offline RL policies are pre-trained on static data before being evaluated in the simulator. All results are averaged over five seeds (11, 15, 17, 19, 23) with standard deviation reported.

\textbf{Evaluation Metrics}. Following~\cite{len}, we also measure user long-term satisfaction in the interaction recommendation environment by \emph{Interaction Length} ($Len$), \emph{Cumulative Reward} ($R_{\mathrm{cum}}$), and \emph{Single-round Reward} ($R_{\mathrm{each}}$), and fairness is measured by \textit{Absolute Difference} (AD)~\cite{ad}, which is a widely used metric for fairness. For implementation reproducibility, key hyperparameters and code availability are discussed in the Appendix.

\textbf{Baselines.} 
To assess the effectiveness of LHRL, we compare it against two categories of representative approaches. (1) \textit{Reinforcement learning-based recommenders}: CQL~\cite{cql}, BCQ~\cite{bcq}, DORL~\cite{dorl}, SQN~\cite{sqn}, DDPG~\cite{ddpg}, and TD3~\cite{td3}, which learn policies from sequential user-system interactions but do not explicitly consider fairness. (2) \textit{Fairness-aware interactive recommenders}: MOFIR~\cite{mofir}, SAC4IR~\cite{sac}, DNAIR~\cite{dna}, FCPO~\cite{fcpo}, and HER4IF~\cite{her}, which incorporate fairness objectives into the RL optimization process.

To ensure a fair comparison, we apply grid search to identify the optimal hyperparameter configurations for each baseline. Detailed parameter settings are provided in the Appendix.

\begin{table*}[!t]
\centering
\resizebox{\textwidth}{!}{%
\begin{tabular}{lcccccccccccc}
\toprule
\multirow{2}{*}{Methods} &
\multicolumn{4}{c}{KuaiRec} & &
\multicolumn{4}{c}{KuaiRand} \\
\cmidrule(lr){2-5} \cmidrule(lr){7-10}
& $Len$ & $R_{\text{each}}$ & $R_{\text{cum}}$ & AD & &
$Len$ & $R_{\text{each}}$ & $R_{\text{cum}}$ & AD \\
\midrule
DORL     & 8.680$\pm$0.204 & 0.575$\pm$0.057 & 4.876$\pm$0.559 & 0.660$\pm$0.075 &&8.960$\pm$0.136 & 0.658$\pm$0.052 & 5.767$\pm$0.565 & 0.947$\pm$0.015 \\
CQL      & 9.167$\pm$0.094 & 0.635$\pm$0.004 & 5.713$\pm$0.092 & 0.684$\pm$0.029 && 9.080$\pm$0.040 & 0.705$\pm$0.008 & 6.268$\pm$0.134 & 0.839$\pm$0.027 \\
BCQ      & 13.320$\pm$0.360 & 0.730$\pm$0.009 & 9.221$\pm$0.331 & 0.018$\pm$0.002 && 10.040$\pm$0.950 & 0.735$\pm$0.011 & 7.124$\pm$0.733 & 0.473$\pm$0.178 \\
\midrule
SQN      & 9.140$\pm$0.049 & 0.631$\pm$0.005 & 5.577$\pm$0.062 & 0.706$\pm$0.004 && 9.140$\pm$0.049 & 0.703$\pm$0.002 & 6.236$\pm$0.034 & 0.831$\pm$0.002 \\
TD3      & 14.680$\pm$2.977 & 0.807$\pm$0.022 & 11.906$\pm$2.822 & 0.140$\pm$0.168 && 10.460$\pm$0.852 & 0.755$\pm$0.010 & 7.147$\pm$0.392 & 0.367$\pm$0.089 \\
DDPG     & 14.200$\pm$2.322 & 0.844$\pm$0.032 & 11.145$\pm$2.739 & 0.186$\pm$0.093 && 11.560$\pm$0.948 & 0.779$\pm$0.011 & 8.379$\pm$0.917 & 0.330$\pm$0.083 \\
\midrule
MOFIR    & 8.720$\pm$0.232 & 0.547$\pm$0.060 & 4.674$\pm$0.597 & 0.755$\pm$0.016 && 9.060$\pm$0.174 & 0.669$\pm$0.033 & 5.936$\pm$0.361 & 0.862$\pm$0.074 \\
FCPO     & 9.880$\pm$0.926  & 0.719$\pm$0.058 & 6.829$\pm$1.120 & 0.609$\pm$0.114 && 13.960$\pm$3.969 & 0.786$\pm$0.053 & 11.023$\pm$3.714 & 0.312$\pm$0.383 \\
DNAIR    & 11.020$\pm$1.155 & 0.714$\pm$0.013 & 7.445$\pm$0.855 & 0.441$\pm$0.166 && 9.300$\pm$0.089 & 0.756$\pm$0.011 & 6.895$\pm$0.132 & 0.741$\pm$0.041 \\
SAC4IR   & 14.960$\pm$0.796 & 0.788$\pm$0.015 & 11.412$\pm$1.090 & 0.012$\pm$0.002 && 9.160$\pm$0.049 & 0.751$\pm$0.007 & 6.779$\pm$0.079 & 0.735$\pm$0.018 \\
HER4IF   & 16.075$\pm$0.238 & 0.886$\pm$0.007 & 13.802$\pm$0.221 & 0.010$\pm$0.002 && 13.880$\pm$3.269 & 0.829$\pm$0.018 & 11.126$\pm$2.815 & 0.220$\pm$0.301 \\
\midrule
\textbf{LHRL (Ours)} & \textbf{17.760$\pm$0.723} & \textbf{0.895$\pm$0.008} & \textbf{15.613$\pm$0.863} & \textbf{0.009$\pm$0.002} && \textbf{15.360$\pm$2.951} & \textbf{0.841$\pm$0.014} & \textbf{12.655$\pm$2.648} & \textbf{0.184$\pm$0.267} \\
\textbf{\%improv.} & \textbf{10.48\%} & \textbf{1.02\%} & \textbf{13.12\%} & \textbf{10.00\%} && \textbf{10.03\%} & \textbf{1.45\%} & \textbf{13.74\%} & \textbf{16.36\%} \\
\bottomrule
\end{tabular}%
}
\caption{Average results of all methods in two environments (Bold: best).}
\label{tab1:performance}
\end{table*}

\subsection{Overall Performance Comparison (RQ1)}
To assess LHRL's effectiveness in harmonizing long-term user satisfaction and fairness, we conduct comprehensive comparisons against 11 state-of-the-art baselines across two datasets. Results in Table~\ref{tab1:performance} reveal the following findings: 1) on the KuaiRec dataset, LHRL achieves the highest scores in all accuracy metrics, with a Len of 17.760 and a $R_{cum}$ of 15.613. This represents a substantial 10.5\% and 13.1\% improvement in long-term user engagement over the strongest baseline. More importantly, this improvement in user satisfaction is achieved while maintaining better exposure fairness, with an AD score of 0.009. This demonstrates a rare synergy, as most methods struggle with a trade-off between user satisfaction and fairness. 2) LHRL's superiority is even more pronounced. It boosts $R_{cum}$ to 12.655, a 13.7\% increase over the runner-up HER4IF, while simultaneously achieving the lowest possible AD score of 0.184, drastically outperforming all other baselines in fairness. These results robustly demonstrate LHRL's ability to harmonize the dual objectives of user engagement and ecosystem fairness.

\subsection{Ablation Study(RQ2)}

\begin{table*}[t]
\centering
\resizebox{\textwidth}{!}{%
\begin{tabular}{lcccccccc}
\toprule
\multirow{2}{*}{Methods} &
\multicolumn{4}{c}{KuaiRec} & 
\multicolumn{4}{c}{KuaiRand} \\
\cmidrule(lr){2-5} \cmidrule(lr){6-9}
& Len & $R_{each}$ & $R_{cum}$ & AD & Len & $R_{each}$ & $R_{cum}$ & AD \\
\midrule
LHRL-w/o A & 10.320$\pm$1.234 & 0.810$\pm$0.009 & 8.460$\pm$1.004 & 0.462$\pm$0.333 & 9.140$\pm$0.403 & 0.759$\pm$0.017 & 6.849$\pm$0.280 & 0.821$\pm$0.154 \\
LHRL-w/o H & 10.700$\pm$1.154 & 0.805$\pm$0.019 & 8.700$\pm$1.122 & 0.330$\pm$0.165 & 10.480$\pm$2.767 & 0.770$\pm$0.043 & 8.262$\pm$2.607 & 0.617$\pm$0.327 \\
LHRL-w/o L & 15.300$\pm$0.477 & 0.864$\pm$0.002 & 13.226$\pm$0.653 & 0.037$\pm$0.014 & 12.340$\pm$2.735 & 0.806$\pm$0.021 & 9.712$\pm$2.330 & 0.459$\pm$0.433 \\
LHRL & \textbf{17.760$\pm$0.723} & \textbf{0.895$\pm$0.008} & \textbf{15.613$\pm$0.863} & \textbf{0.009$\pm$0.002} & \textbf{15.360$\pm$2.951} & \textbf{0.841$\pm$0.014} & \textbf{12.655$\pm$2.648} & \textbf{0.184$\pm$0.267} \\
\bottomrule
\end{tabular}%
}
\caption{Ablation study of LHRL.}
\label{tab:ablation}
\end{table*}
To validate the necessity of our two key design choices—lifecycle-awareness and the hierarchical structure, we compare our full model against three variants: 1) LHRL-w/o L: The hierarchical model without lifecycle-aware rewards. 2) LHRL-w/o H: A single-level RL agent that uses lifecycle-aware rewards. 3) LHRL-w/o A: A standard flat RL agent without any of our contributions.

The results are shown in Table~\ref{tab:ablation}. We can find that: 1) Compared to LHRL-w/o L, we observe that removing the lifecycle rewards leads to a noticeable drop in both Interaction Length and Cumulative Reward. This confirms that guiding the agent with lifecycle information is critical for maximizing long-term user engagement. 2) Compared to LHRL-w/o H, it reveals the importance of the hierarchy. While the flat agent initially performs well, its performance is unstable and collapses after approximately 10,000 steps, particularly in long-term metrics (Len and $R_{\mathrm{cum}}$). This demonstrates that a non-hierarchical approach struggles to balance short-term accuracy with long-term fairness constraints, resulting in poor long-term stability and generalization, leading to policy degradation. 3) The LHRL-w/o A variant performs the poorest across all metrics, highlighting that the combination of lifecycle-awareness and hierarchical control works synergistically to achieve the robust, state-of-the-art performance of our full model. Thus, the experiment results demonstrate that both components are essential.

\subsection{Analysis of the PhaseFormer Module (RQ3)}
\begin{figure}[!t]
    \centering
    \begin{subfigure}{0.48\linewidth}
        \includegraphics[width=\textwidth]{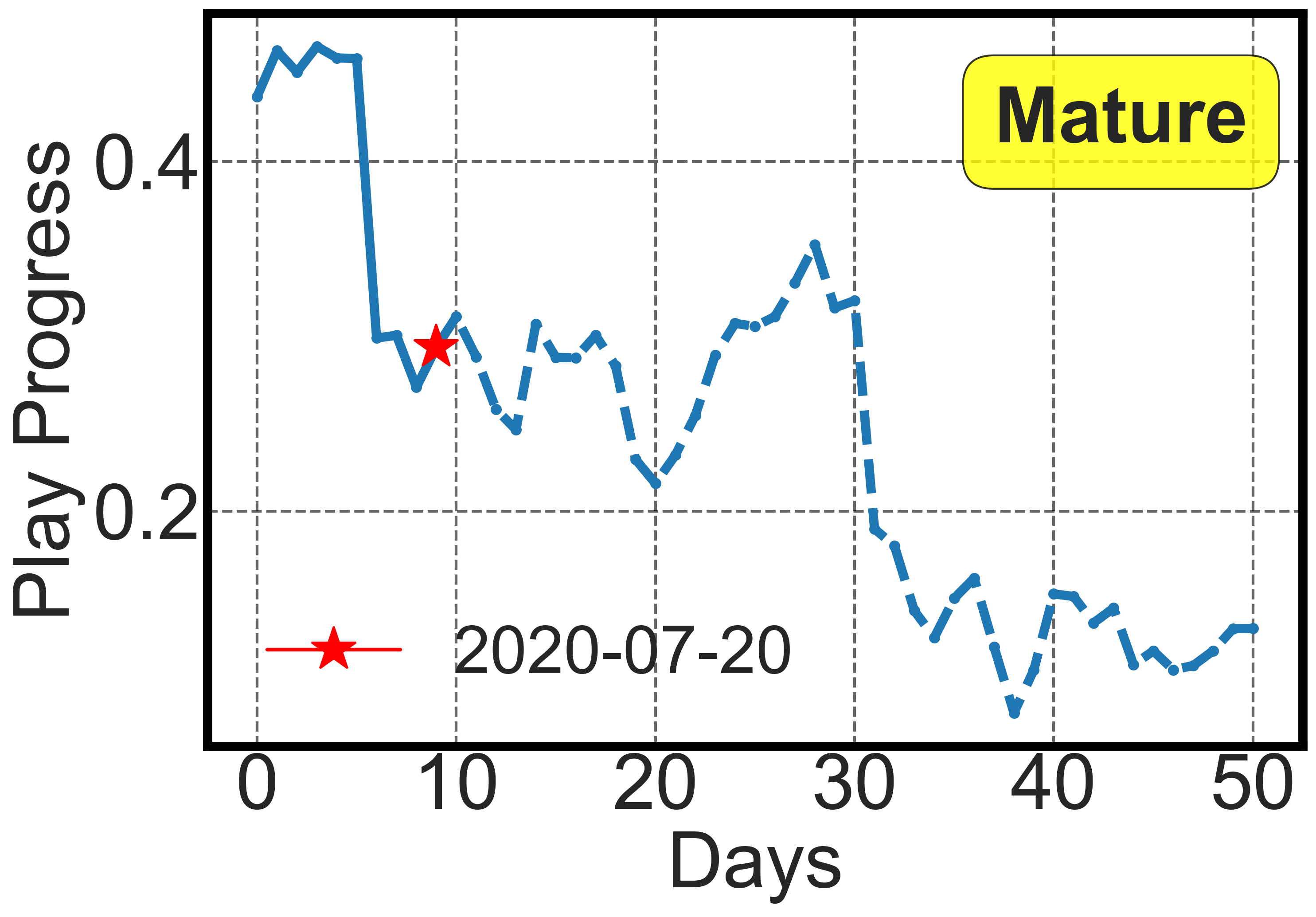}
    \end{subfigure}
    \begin{subfigure}{0.48\linewidth}
        \includegraphics[width=\textwidth]{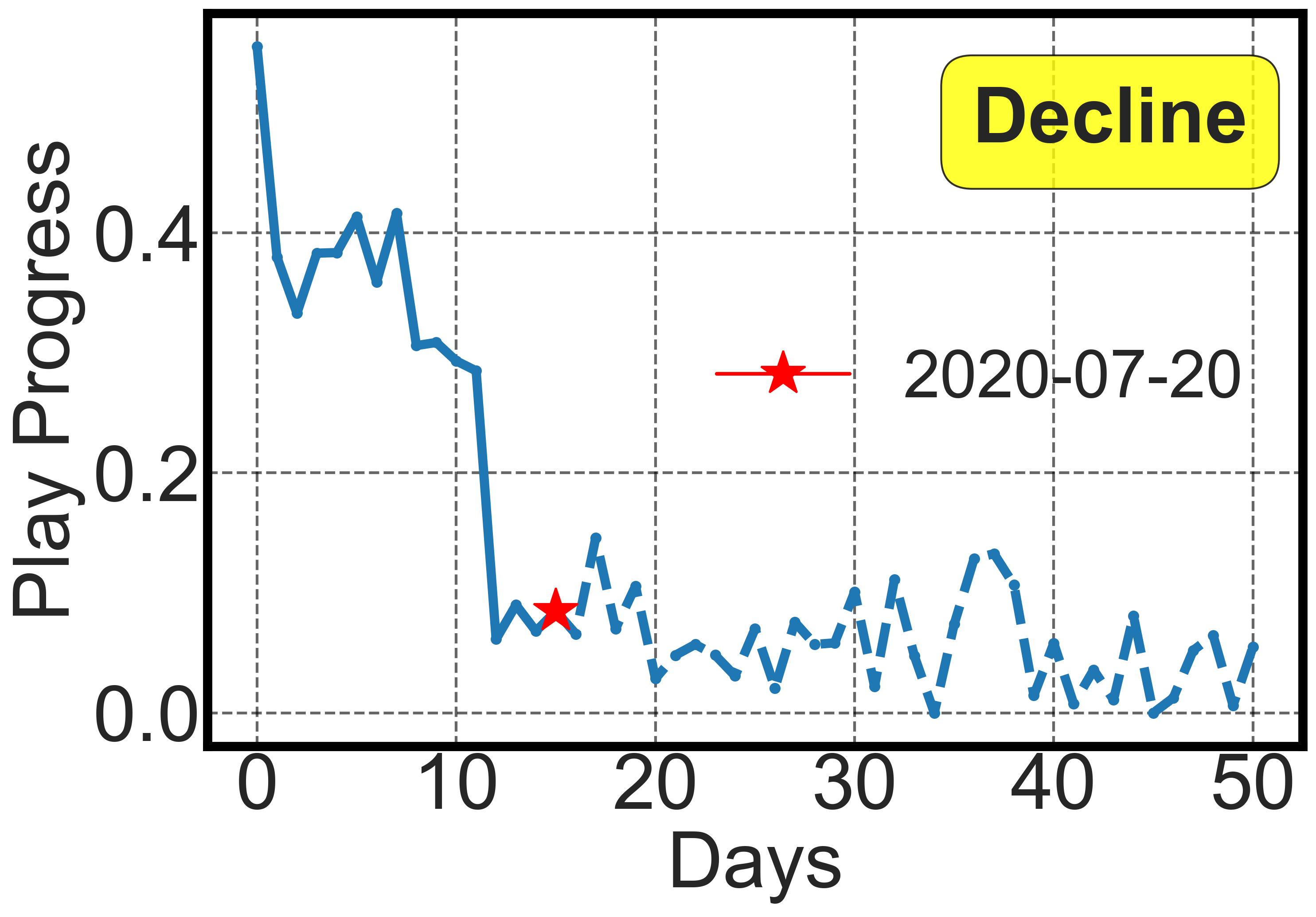}
    \end{subfigure}
    \caption{Case studies of the PhaseFormer module's predictions. }
    \label{fig:case_study}
\end{figure}
To further validate the effectiveness of our PhaseFormer module, we provide a qualitative analysis of its predictions on individual items. Figure~\ref{fig:case_study} presents the real-world play progress trajectories of two videos, randomly sampled from the dataset, along with the lifecycle stage predicted by our module at a specific point in time (2020-07-20). Crucially, this prediction is made using only the historical data available up to that point, simulating how the model would perform in a real-time system with incomplete future information. These examples demonstrate that our PhaseFormer module learns to make semantically meaningful and accurate judgments about an item's current value based on its historical trajectory. This fine-grained understanding serves as a reliable foundation for the high-level agent to dynamically control the recommendation strategy, leading to the robust performance gains documented in our quantitative experiments.

\subsection{Generalizability of Lifecycle-Aware Rewards (RQ4)}

\begin{table*}[t]
\centering
\resizebox{\textwidth}{!}{%
\begin{tabular}{lcccccccccccc}
\toprule
\multirow{2}{*}{Methods} &
\multicolumn{4}{c}{KuaiRec} & &
\multicolumn{4}{c}{KuaiRand} \\
\cmidrule(lr){2-5} \cmidrule(lr){7-10}
& $Len$ & $R_{\text{each}}$ & $R_{\text{cum}}$ & AD & &
$Len$ & $R_{\text{each}}$ & $R_{\text{cum}}$ & AD \\
\midrule
SAC4IR        & 14.960$\pm$0.796 & 0.788$\pm$0.015 & 11.412$\pm$1.090 & 0.012$\pm$0.002 && 9.160$\pm$0.049 & 0.751$\pm$0.007 & 6.779$\pm$0.079 & 0.735$\pm$0.018 \\
SAC4IR+life   & 15.400$\pm$0.424 & 0.791$\pm$0.014 & 12.346$\pm$0.277 & 0.013$\pm$0.002 && 9.200$\pm$0.082 & 0.758$\pm$0.005 & 6.768$\pm$0.045 & 0.666$\pm$0.025 \\
\midrule
DNAIR         & 11.020$\pm$1.155 & 0.714$\pm$0.013 & 7.445$\pm$0.855 & 0.441$\pm$0.166 && 9.300$\pm$0.089 & 0.756$\pm$0.011 & 6.895$\pm$0.132 & 0.741$\pm$0.041 \\
DNAIR+life    & 11.200$\pm$1.400 & 0.739$\pm$0.027 & 8.078$\pm$1.465 & 0.412$\pm$0.175 && 9.433$\pm$0.170 & 0.756$\pm$0.011 & 6.894$\pm$0.101 & 0.652$\pm$0.073 \\
\midrule
FCPO          & 9.880$\pm$0.926  & 0.719$\pm$0.058 & 6.829$\pm$1.120 & 0.609$\pm$0.114 && 13.960$\pm$3.969 & 0.786$\pm$0.053 & 11.023$\pm$3.714 & 0.312$\pm$0.383 \\
FCPO+life     & 10.600$\pm$0.963 & 0.731$\pm$0.007 & 7.238$\pm$0.423 & 0.431$\pm$0.096 && 14.767$\pm$3.441 & 0.800$\pm$0.045 & 11.650$\pm$3.526 & 0.252$\pm$0.356 \\
\bottomrule
\end{tabular}%
}
\caption{Performance comparison of life-enhanced methods on two environments.}
\label{tab:life_comparison}
\end{table*}

To demonstrate that our proposed lifecycle-aware reward is a generalizable concept, we integrated it into several existing fairness-aware RL baselines. Table \ref{tab:life_comparison} shows the results of adding our lifecycle module to SAC4IR, DNAIR, and FCPO. The results show consistent performance improvements. For instance, SAC4IR+life improves $R_{cum}$ by 8.2\% on KuaiRec, and FCPO+life improves $R_{cum}$ by 5.7\% on KuaiRand. This demonstrates that lifecycle modeling is a portable and effective principle for enhancing fairness-aware recommendation, and its benefits are not limited to our specific HRL architecture. A full analysis of hyperparameter sensitivity is left as future work.

\subsection{Analysis of Learned Recommendation Strategy (RQ5)}
\begin{figure}[t]
    \centering
    \begin{subfigure}{0.48\linewidth}
        \includegraphics[width=\textwidth]{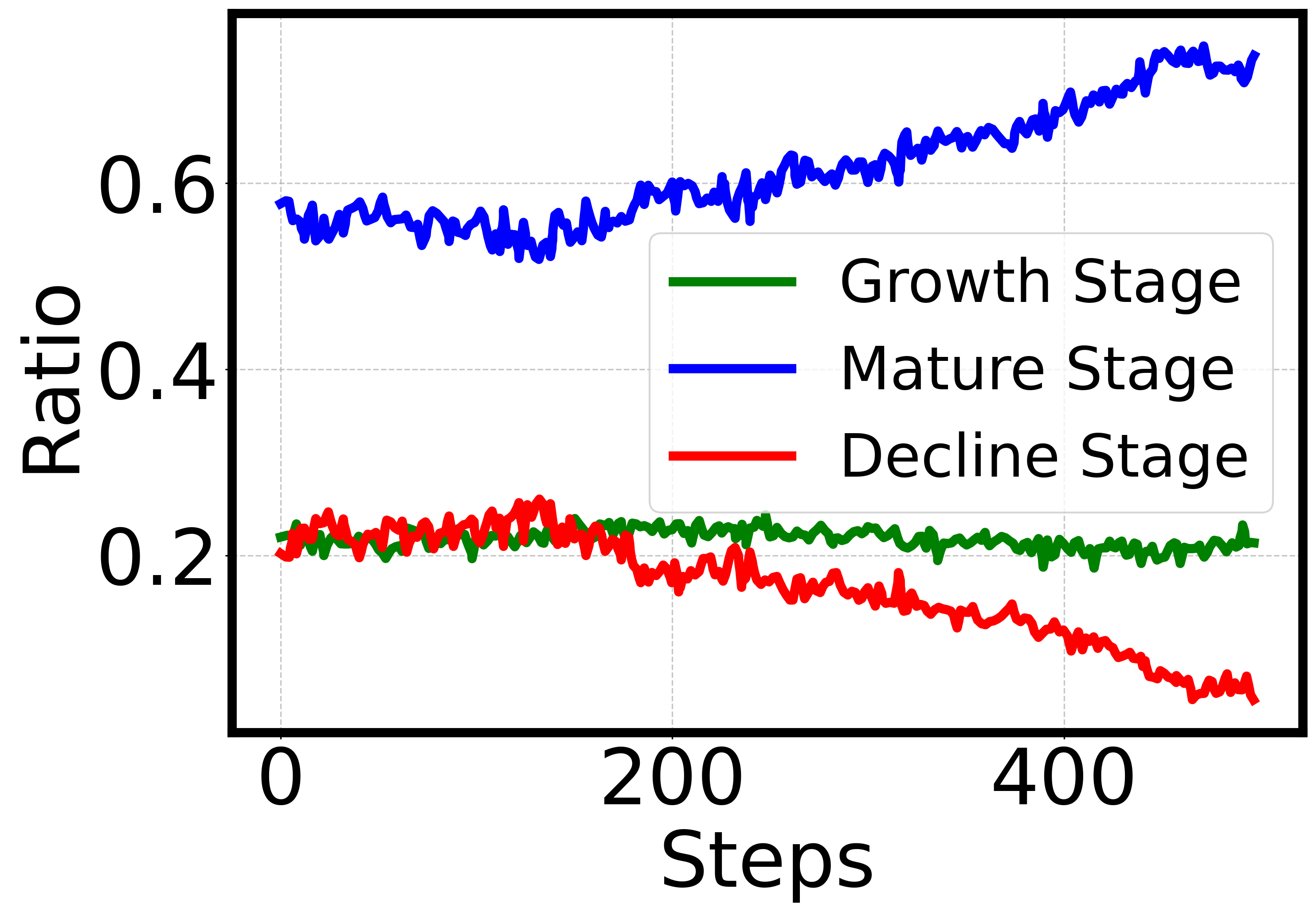}
        \caption{KuaiRec}
    \end{subfigure}
    \begin{subfigure}{0.48\linewidth}
        \includegraphics[width=\textwidth]{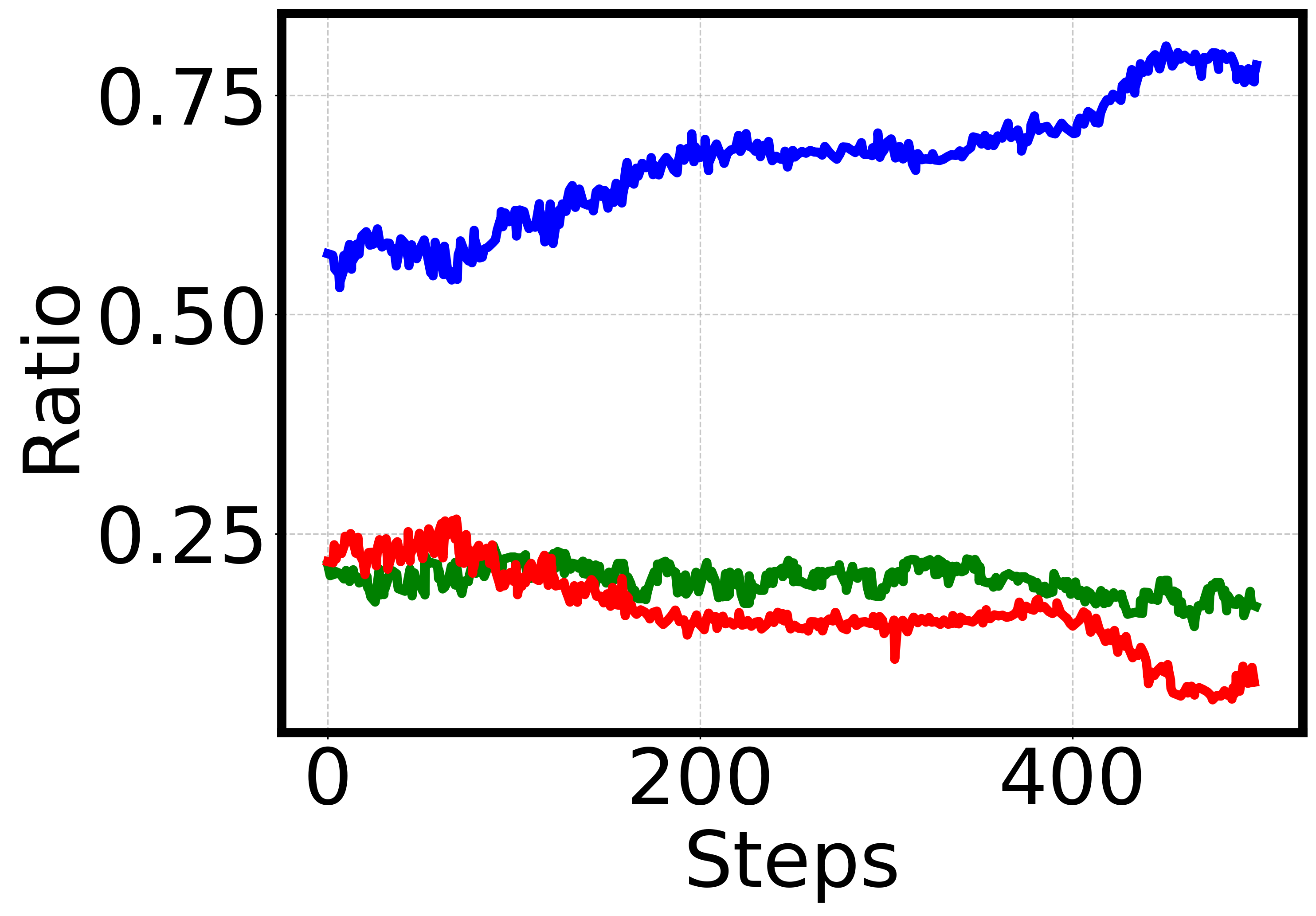}
        \caption{KuaiRand}
    \end{subfigure}
    \caption{Proportion of videos from different lifecycle stages in the recommendation list over time. }
    \label{fig:ratio_plots}
\end{figure}
To understand how LHRL achieves its superior performance, we analyzed the composition of its recommendation lists over the training process. Figure \ref{fig:ratio_plots} illustrates the proportion of items from each lifecycle stage (Growth, Mature, and Decline) that the agent chooses to recommend. A clear and consistent strategy emerges on both datasets. As training progresses, the LHRL agent learns to increase the exposure of Mature Stage items, which have high, stable engagement value, gradually decrease the exposure of Decline Stage items, actively pruning content that is losing user interest, and maintain a stable, low level of exposure for Growth Stage items, allowing for exploration without over-promoting unproven content. This learned behavior is the core mechanism behind LHRL's success. By intelligently reallocating exposure from low-value, declining items to high-value, mature items, the model creates a healthier content ecosystem. This directly translates into the extended user interaction lengths and higher cumulative rewards observed in Table~\ref{tab1:performance}.

\section{Related Work}
Research on fairness in recommendation systems can be broadly categorized into static and dynamic approaches. Static methods typically focus on one-shot recommendation settings, which neglect the temporal dynamics of user behavior and item value. Representative approaches include incorporating fairness-aware regularization terms during model training, leveraging causal inference to disentangle users' true preferences from platform interventions, and applying re-ranking algorithms at the post-processing stage to balance exposure. While these methods enhance the overall exposure of long-tail items, they often exhibit limitations when addressing the long-term evolution of user-item interactions.

To address the challenges posed by the evolving nature of user interests and item popularity in real-world scenarios, recent studies have introduced Reinforcement Learning (RL) to construct dynamic frameworks for fair recommendation. These approaches primarily fall into three categories: (1) Dynamic Exposure Constraint Strategies, where methods\cite{r-19} dynamically adjust the proportion of popular content based on temporal shifts; (2) Debiased Reward Modeling, such as SAC4IR\cite{sac}, which encourages the discovery of non-popular items through a maximum entropy reward function; and (3) Long-term Fairness Optimization, where frameworks like FCPO\cite{fcpo}, DNAIR\cite{dna}, and HER4IF\cite{her} employ multi-objective RL for long-term policy optimization. Despite these advances in modeling long-term fairness, these methods typically overlook the varying exposure requirements of items at different stages of their lifecycle. However, the "Product Lifecycle" theory\cite{levitt1965exploit,draskovic2018product} from marketing science posits that items possess distinct value propositions and levels of user appeal at different stages. Indeed, on platforms such as those for short-form videos, an item's lifecycle often exhibits a phased pattern of "growth-mature-decline." Consequently, employing a uniform exposure strategy is insufficient for achieving true dynamic fairness while simultaneously maintaining user interest. Building on this insight, we are the first to leverage the item lifecycle as a central modulating factor for dynamic fairness adjustment by proposing the LHRL framework. By precisely predicting an item's current stage (growth, mature, or decline) and applying phase-aware exposure strategies, our framework offers a novel and more fine-grained approach to resolving the inherent tension between long-term fairness and user engagement.

\section{Conclusion}
In this paper, we addressed the critical challenge of fairness and accuracy in interactive recommender systems by introducing the item lifecycle as a novel, dynamic control knob. We proposed LHRL, a Lifecycle-aware Hierarchical Reinforcement Learning framework, which uniquely leverages a temporal-aware PhaseFormer module to identify an item's real-time stage (growth, mature, or decline). This lifecycle information guides our hierarchical agent to decouple the optimization of long-term fairness and short-term user engagement, enabling it to dynamically set phase-specific exposure constraints. Extensive experiments on real-world datasets demonstrate that LHRL achieves state-of-the-art performance in both accuracy and fairness, significantly outperforming existing methods. Our findings validate that explicitly modeling item lifecycles is a powerful and effective strategy for building more sustainable and equitable recommendation ecosystems.

\section{Acknowledgments}
This work was supported by the National Natural Science Foundation of China (No. 62476261), in part by the Chongqing Natural Science Foundation Innovation and Development Joint Fund (No. CSTB2025NSCQ-LZX0061), and in part by the Science and Technology Innovation Key R\&D Program of Chongqing (CSTB2023TIAD-STX0031, CSTB2025TIAD-STX0023).

\bibliography{aaai2026}

\begin{thebibliography}{25}
\providecommand{\natexlab}[1]{#1}

\bibitem[{Abdollahpouri, Burke, and Mobasher(2019)}]{abdollahpouri2019managing}
Abdollahpouri, H.; Burke, R.; and Mobasher, B. 2019.
\newblock Managing popularity bias in recommender systems with personalized re-ranking.
\newblock \emph{arXiv preprint arXiv:1901.07555}.

\bibitem[{Draskovic, Markovic, and Znidar(2018)}]{draskovic2018product}
Draskovic, N.; Markovic, M.; and Znidar, K. 2018.
\newblock Product lifecycle strategies in digital world.
\newblock \emph{International journal of management cases}, 20(3): 29--43.

\bibitem[{Fujimoto, Hoof, and Meger(2018)}]{td3}
Fujimoto, S.; Hoof, H.; and Meger, D. 2018.
\newblock Addressing function approximation error in actor-critic methods.
\newblock In \emph{International conference on machine learning}, 1587--1596. PMLR.

\bibitem[{Fujimoto, Meger, and Precup(2019)}]{bcq}
Fujimoto, S.; Meger, D.; and Precup, D. 2019.
\newblock Off-policy deep reinforcement learning without exploration.
\newblock In \emph{International conference on machine learning}, 2052--2062. PMLR.

\bibitem[{Gao et~al.(2023)Gao, Huang, Chen, Zhang, Li, Jiang, Wang, Zhang, and He}]{dorl}
Gao, C.; Huang, K.; Chen, J.; Zhang, Y.; Li, B.; Jiang, P.; Wang, S.; Zhang, Z.; and He, X. 2023.
\newblock Alleviating matthew effect of offline reinforcement learning in interactive recommendation.
\newblock In \emph{Proceedings of the 46th International ACM SIGIR Conference on Research and Development in Information Retrieval}, 238--248.

\bibitem[{Gao et~al.(2022{\natexlab{a}})Gao, Li, Lei, Li, Jiang, Chen, He, Mao, and Chua}]{rec}
Gao, C.; Li, S.; Lei, W.; Li, B.; Jiang, P.; Chen, J.; He, X.; Mao, J.; and Chua, T.-S. 2022{\natexlab{a}}.
\newblock KuaiRec: A fully-observed dataset for recommender systems.
\newblock \emph{arXiv preprint arXiv:2202.10842}.

\bibitem[{Gao et~al.(2022{\natexlab{b}})Gao, Li, Zhang, Chen, Li, Lei, Jiang, and He}]{rand}
Gao, C.; Li, S.; Zhang, Y.; Chen, J.; Li, B.; Lei, W.; Jiang, P.; and He, X. 2022{\natexlab{b}}.
\newblock Kuairand: an unbiased sequential recommendation dataset with randomly exposed videos.
\newblock In \emph{Proceedings of the 31st ACM International Conference on Information \& Knowledge Management}, 3953--3957.

\bibitem[{Ge et~al.(2021)Ge, Liu, Gao, Xian, Li, Zhao, Pei, Sun, Ge, Ou et~al.}]{fcpo}
Ge, Y.; Liu, S.; Gao, R.; Xian, Y.; Li, Y.; Zhao, X.; Pei, C.; Sun, F.; Ge, J.; Ou, W.; et~al. 2021.
\newblock Towards long-term fairness in recommendation.
\newblock In \emph{Proceedings of the 14th ACM international conference on web search and data mining}, 445--453.

\bibitem[{Ge et~al.(2022)Ge, Zhao, Yu, Paul, Hu, Hsieh, and Zhang}]{mofir}
Ge, Y.; Zhao, X.; Yu, L.; Paul, S.; Hu, D.; Hsieh, C.-C.; and Zhang, Y. 2022.
\newblock Toward pareto efficient fairness-utility trade-off in recommendation through reinforcement learning.
\newblock In \emph{Proceedings of the fifteenth ACM international conference on web search and data mining}, 316--324.

\bibitem[{Konstantinov and Lampert(2021)}]{konstantinov2021fairness}
Konstantinov, N.; and Lampert, C.~H. 2021.
\newblock Fairness through regularization for learning to rank.
\newblock \emph{arXiv preprint arXiv:2102.05996}.

\bibitem[{Kumar et~al.(2020)Kumar, Zhou, Tucker, and Levine}]{cql}
Kumar, A.; Zhou, A.; Tucker, G.; and Levine, S. 2020.
\newblock Conservative q-learning for offline reinforcement learning.
\newblock \emph{Advances in Neural Information Processing Systems}, 33: 1179--1191.

\bibitem[{Levitt et~al.(1965)}]{levitt1965exploit}
Levitt, T.; et~al. 1965.
\newblock \emph{Exploit the product life cycle}, volume~43.
\newblock Graduate School of Business Administration, Harvard University Cambridge, MA~….

\bibitem[{Lillicrap(2015)}]{ddpg}
Lillicrap, T. 2015.
\newblock Continuous control with deep reinforcement learning.
\newblock \emph{arXiv preprint arXiv:1509.02971}.

\bibitem[{Liu et~al.(2021)Liu, Liu, Tang, Liao, Chen, and Heng}]{r-19}
Liu, W.; Liu, F.; Tang, R.; Liao, B.; Chen, G.; and Heng, P.~A. 2021.
\newblock Balancing accuracy and fairness for interactive recommendation with reinforcement learning.
\newblock \emph{arXiv preprint arXiv:2106.13386}.

\bibitem[{Morik et~al.(2020)Morik, Singh, Hong, and Joachims}]{r-23}
Morik, M.; Singh, A.; Hong, J.; and Joachims, T. 2020.
\newblock Controlling fairness and bias in dynamic learning-to-rank.
\newblock In \emph{Proceedings of the 43rd international ACM SIGIR conference on research and development in information retrieval}, 429--438.

\bibitem[{Schulman et~al.(2017)Schulman, Wolski, Dhariwal, Radford, and Klimov}]{ppo}
Schulman, J.; Wolski, F.; Dhariwal, P.; Radford, A.; and Klimov, O. 2017.
\newblock Proximal policy optimization algorithms.
\newblock \emph{arXiv preprint arXiv:1707.06347}.

\bibitem[{Shi et~al.(2024{\natexlab{a}})Shi, He, Zhang, Gao, Li, Zhang, Wang, and Feng}]{len}
Shi, W.; He, X.; Zhang, Y.; Gao, C.; Li, X.; Zhang, J.; Wang, Q.; and Feng, F. 2024{\natexlab{a}}.
\newblock Large language models are learnable planners for long-term recommendation.
\newblock In \emph{Proceedings of the 47th International ACM SIGIR Conference on Research and Development in Information Retrieval}, 1893--1903.

\bibitem[{Shi et~al.(2024{\natexlab{b}})Shi, Liu, Xie, Bai, and Shang}]{sac}
Shi, X.; Liu, Q.; Xie, H.; Bai, Y.; and Shang, M. 2024{\natexlab{b}}.
\newblock Maximum Entropy Policy for Long-Term Fairness in Interactive Recommender Systems.
\newblock \emph{IEEE Transactions on Services Computing}.

\bibitem[{Shi et~al.(2023)Shi, Liu, Xie, Wu, Peng, Shang, and Lian}]{dna}
Shi, X.; Liu, Q.; Xie, H.; Wu, D.; Peng, B.; Shang, M.; and Lian, D. 2023.
\newblock Relieving popularity bias in interactive recommendation: A diversity-novelty-aware reinforcement learning approach.
\newblock \emph{ACM Transactions on Information Systems}, 42(2): 1--30.

\bibitem[{Wang et~al.(2023)Wang, Ma, Zhang, Liu, and Ma}]{ad}
Wang, Y.; Ma, W.; Zhang, M.; Liu, Y.; and Ma, S. 2023.
\newblock A survey on the fairness of recommender systems.
\newblock \emph{ACM Transactions on Information Systems}, 41(3): 1--43.

\bibitem[{Wei et~al.(2021)Wei, Feng, Chen, Wu, Yi, and He}]{wei2021model}
Wei, T.; Feng, F.; Chen, J.; Wu, Z.; Yi, J.; and He, X. 2021.
\newblock Model-agnostic counterfactual reasoning for eliminating popularity bias in recommender system.
\newblock In \emph{Proceedings of the 27th ACM SIGKDD conference on knowledge discovery \& data mining}, 1791--1800.

\bibitem[{Xia et~al.(2024)Xia, Shi, Xie, Liu, and Shang}]{her}
Xia, C.; Shi, X.; Xie, H.; Liu, Q.; and Shang, M. 2024.
\newblock Hierarchical Reinforcement Learning for Long-term Fairness in Interactive Recommendation.
\newblock In \emph{2024 IEEE International Conference on Web Services (ICWS)}, 300--309. IEEE.

\bibitem[{Xin et~al.(2020)Xin, Karatzoglou, Arapakis, and Jose}]{sqn}
Xin, X.; Karatzoglou, A.; Arapakis, I.; and Jose, J.~M. 2020.
\newblock Self-supervised reinforcement learning for recommender systems.
\newblock In \emph{Proceedings of the 43rd International ACM SIGIR conference on research and development in Information Retrieval}, 931--940.

\bibitem[{Zheng et~al.(2021)Zheng, Gao, Li, He, Li, and Jin}]{zheng2021disentangling}
Zheng, Y.; Gao, C.; Li, X.; He, X.; Li, Y.; and Jin, D. 2021.
\newblock Disentangling user interest and conformity for recommendation with causal embedding.
\newblock In \emph{Proceedings of the web conference 2021}, 2980--2991.

\bibitem[{Zhu et~al.(2021)Zhu, He, Zhao, Zhang, Wang, and Caverlee}]{zhu2021popularity}
Zhu, Z.; He, Y.; Zhao, X.; Zhang, Y.; Wang, J.; and Caverlee, J. 2021.
\newblock Popularity-opportunity bias in collaborative filtering.
\newblock In \emph{Proceedings of the 14th ACM international conference on web search and data mining}, 85--93.

\end{thebibliography}

\end{document}